\documentclass{article}

\usepackage[final]{neurips_2024_calm}

\usepackage[utf8]{inputenc} 
\usepackage[T1]{fontenc}    
\usepackage{hyperref}       
\usepackage{url}            
\usepackage{booktabs}       
\usepackage{amsfonts}       
\usepackage{nicefrac}       
\usepackage{microtype}      
\usepackage{xcolor}         
\usepackage{subfig}
\usepackage{caption}
\usepackage{multirow}
\usepackage{graphicx}
\usepackage{amsmath} 
\usepackage{tikz} 

\newcommand*\circled[1]{\tikz[baseline=(char.base)]{
    \node[shape=circle,draw,inner sep=0pt, minimum size=9pt] (char) {#1};}}

\title{Causal Reasoning in Large Language Models: \\
A Knowledge Graph Approach}

\author{%
  Yejin Kim \\
  GraphLab \\
  George Washington University\\
  Washington, DC, USA \\
  \texttt{yejinjenny@gwu.edu} \\
  \And
  Eojin Kang \\
  AI Convergence \\
  Hankuk University of Foreign Studies \\
  Seoul, Korea \\
  \texttt{eojinkang@hufs.ac.kr} \\
  \And
  Juae Kim$^{*}$ \\
  AI Convergence \\
  Hankuk University of Foreign Studies \\
  Seoul, Korea \\
  \texttt{juaekim@hufs.ac.kr} \\
  \And
  H. Howie Huang\thanks{These authors are corresponding authors.} \\
  GraphLab \\
  George Washington University \\
  Washington, DC, USA \\
  \texttt{howie@gwu.edu} \\
}

\begin{document}

\maketitle

\begin{abstract}
Large language models (LLMs) typically improve performance by either retrieving semantically similar information, or enhancing reasoning abilities through structured prompts like chain-of-thought. While both strategies are considered crucial, it remains unclear which has a greater impact on model performance or whether a combination of both is necessary. This paper answers this question by proposing a knowledge graph (KG)-based random-walk reasoning approach that leverages causal relationships. We conduct experiments on the commonsense question answering task that is based on a KG. The KG inherently provides both relevant information, such as related entity keywords, and a reasoning structure through the connections between nodes. Experimental results show that the proposed KG-based random-walk reasoning method improves the reasoning ability and performance of LLMs. Interestingly, incorporating three seemingly irrelevant sentences into the query using KG-based random-walk reasoning enhances LLM performance, contrary to conventional wisdom. These findings suggest that integrating causal structures into prompts can significantly improve reasoning capabilities, providing new insights into the role of causality in optimizing LLM performance.

\end{abstract}

\section{Introduction}

Large language models (LLMs) have demonstrated significant advancements in natural language processing tasks through two primary approaches: providing auxiliary information via retrieval and enhancing reasoning abilities within prompts. Retrieval-augmented generation (RAG) \citep{lewis2020retrieval} is designed to provide information relevant to the given context or query. This relevant information is identified using embedding similarity searches and then integrated into the LLM’s prompt, thereby enhancing the accuracy, relevance recency of the generated response. Recent studies have shown that RAG can significantly boost performance in tasks such as summarization and question answering (QA) \citep{rh:conf/aclnut/GuYW19, rh_DBLP:conf/emnlp/0001PCKW21, rh_DBLP:conf/acl/Komeili0W22}. However, this approach primarily focuses on retrieving directly related information, raising the question of whether relying solely on such relevant information is sufficient for achieving optimal LLM performance.

Another approach to improving the quality of LLM responses is to enhance their reasoning abilities \citep{raesoning_jain2023mechanistically, raesoning_DBLP:conf/acl/SuzgunSSGTCCLCZ23, DBLP:conf/nips/Wei0SBIXCLZ22, DBLP:conf/nips/KojimaGRMI22}. Reasoning capabilities enable LLMs to go beyond surface-level information and make logical connections between concepts, thereby enhancing their ability to handle more complex queries \citep{raesoning_jain2023mechanistically, raesoning_DBLP:conf/acl/SuzgunSSGTCCLCZ23, DBLP:conf/nips/Wei0SBIXCLZ22, DBLP:conf/nips/KojimaGRMI22}. Structured reasoning techniques, such as knowledge graph (KG)-based methods, offer an effective alternative by allowing LLMs to utilize structured knowledge and causal relationships for deeper reasoning \citep{ling2023knowledge, yao2023beyond, DBLP:series/tanlp/SpeerH13}. This capability makes KGs a promising resource for comprehending concepts, applying logical reasoning, and refining or validating the model’s understanding using existing knowledge \citep{wang2023knowledge, yao2023beyond}.

\begin{figure*}
\centering 
\subfloat[KG example of QA]{\includegraphics[width=.43\linewidth] 
{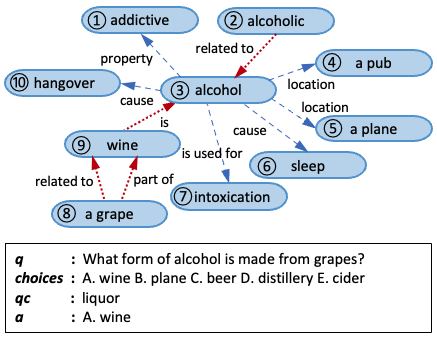}}\hspace{5pt}
\subfloat[KG-Based Reasoning]{\includegraphics[width=.55\linewidth]{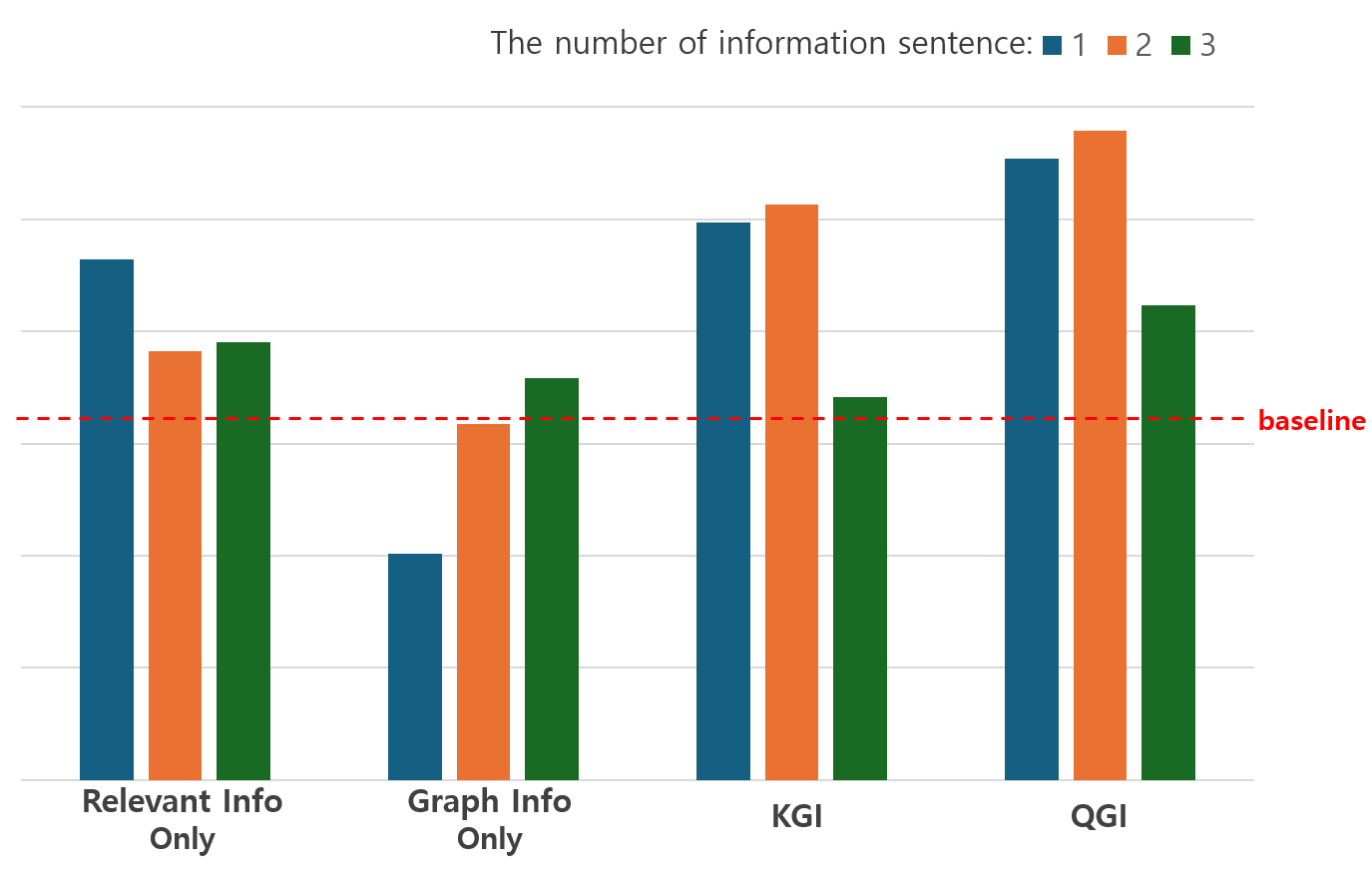}}
\caption{(a) Illustration of a KG structure and an example of CommonsenseQA \citep{talmor2018commonsenseqa}. At the bottom, the question and its concept are represented as $q$ and $qc$ respectively, while the answer is denoted as $a$. (b) Performance comparison on commonsense QA. The dotted line (baseline) represents the performance when no additional context is provided through the prompt. The three bars represent the number of sentences provided as context in the prompt.} 
\label{fig:enter-label}
\end{figure*}

In this paper, we aim to investigate the relative contributions of semantic information retrieval and causal reasoning to the LLMs. To  this end, we propose a KG-based random-walk reasoning approach that navigates paths of interconnected nodes and edges to uncover causal relationships and extract contextual information, thereby enhancing the reasoning capabilities of LLMs. We use the CommonsenseQA \citep{talmor2018commonsenseqa} dataset, constructed from the KG, ConceptNet \citep{speer2017conceptnet}, as our evaluation benchmark. Figure \ref{fig:enter-label}-(a) shows a KG and an example of CommonsenseQA. This dataset is suitable for our study as it requires both relevant information retrieval and complex reasoning to solve problems based on the structured relationships within the KG. We systematically assess the impact of providing semantically relevant information and causal reasoning by conducting experiments with various settings that control the presence and type of contextual information provided to the LLM.

Specifically, we evaluate the effect of incorporating causal relationships extracted through random-walk reasoning within the ConceptNet graph, measuring how these elements contribute to LLM performance in answering commonsense questions. To comprehensively analyze the contributions of information retrieval and reasoning, we designed the following experimental settings:
\begin{enumerate}
    \item \textbf{Relevant Information Only}: In this setting, the context provided to the LLM consists of information with high embedding similarity to the question, inspired by RAG, without any additional reasoning process.

    \item \textbf{Graph Inference Only}: The context provided offers reasoning ability through a continuous process of graph inference, but the information is entirely unrelated to the question.

    \item \textbf{Keyword Relevant Information + Graph Inference (KGI)}: The information means sentences related to the keywords of the question, also providing reasoning ability through the proposed approach. However, there is no guarantee that the information in the graph is directly related to the content of the question. In other words, reasoning and information are provided together, but it is possible to include irrelevant information to the question.

    \item \textbf{Query Relevant Information + Graph Inference (QGI)}: This setting explores how providing both relevant information related to the question and causal relationship through the KG-based random-walk reasoning can enhance the LLMs.
\end{enumerate}

Figure \ref{fig:enter-label}-(b) provides a summary of the experiments. Compared to the baseline, where no additional context is given, the prompt involving either relevant information or causal reasoning improves performance. Notably, in the "Graph Inference Only" setting, when causal relationships were conveyed through the reasoning of two or three sentences, performance improved even though the content was entirely unrelated to the question. This result indicates that our experimental settings successfully convey reasoning capabilities through causal structures. Furthermore, this finding suggests that, contrary to conventional wisdom, it may be more advantageous to equip LLMs with reasoning abilities, grounded in causality, rather than merely providing semantically related information. A comparison between the "Graph Inference Only" and "KGI" settings shows that information extracted through the KG-based random-walk reasoning method yields better performance than the embedding similarity search-based method in the "Graph Inference Only" setting. This demonstrates that the proposed method, which leverages causal relationships through KG, is more effective for commonsense QA. Lastly, the highest improvement is observed in the "QGI" setting, which combines both relevant information and causal reasoning, aligning with our expectations.

Our contributions are summarized as follows:
\begin{itemize}
    \item We demonstrate the contributions of relevant information and reasoning abilities through experimental comparisons.
    
    \item We propose a novel KG-based random-walk reasoning method to utilize causal relationship.
    
    \item We show that providing reasoning capabilities grounded in causal relationships can lead to performance improvements, even when using seemingly unrelated information.
\end{itemize}

\section{Method}
Our main goal is to investigate KG-based reasoning for the commonsense QA task without further training on LLMs. We first briefly introduce problem formulation. Subsequently, we examine the disparities in prompting procedures between the conventional retrieve-based method and our KG-based reasoning approach.



\subsection{Problem Formulation} \label{3_1} 
 In a multiple-choice format such as Figure \ref{fig:enter-label}-(a), the goal is to predict an answer $a \in Aq$ given a pair of a question and a question concept $(q, qc)$, where $q \in Q$. Any keyword capable of categorizing a question could potentially serve as $qc$. In our research, we use the term question concept, $qc$, to encompass all such keywords. The set of choices denoted as $Aq$, varies with each question, and both questions and answers are represented as variable-length text sequences. KG denoted as $G=(V,E)$, is configured as a heterogeneous graph in general. Within this graph, $V$ is the set of entity nodes, and $E \subseteq V \times R \times V$ denotes the set of edges connecting nodes in $V$, where $R$ constitutes a set of relation types. The node $v_{qc} \in V$ denotes the node in the KG that is most semantically similar to $qc$ which means that $v_{qc}$ captures the essence of $qc$ within the graph structure. Using $v_{qc}$ as a starting point, we can explore $n$-hop neighbors which can gather additional information related to the question, $q$. This process is based on the flow of the graph and allows for effective graph reasoning; it adheres to the directionality of edges within the graph structure. Our graph reasoning process aligns with the inherent structure and semantics of the KG.

\begin{figure}
    \centering
    \includegraphics[width=1.0\textwidth]{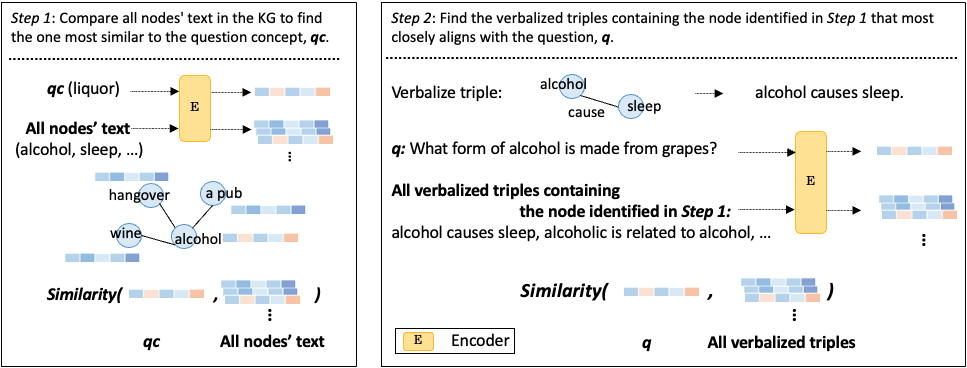}
    \caption{Detailed process of prompting through the KG-based reasoning.}
    \label{fig:model}
\end{figure}

\subsection{Retrieval-Based Prompting}  \label{3_2}
A conventional retrieval-based prompting such as RAG follows:
{\small
\begin{equation}\label{eq1}
\prod_{i}^{N}\sum_{d \in top\text{-}k(p(\cdot|x))}P_\eta(d|x)P_\theta(y_i|x,d,y_{1:i-1})
\end{equation}
}
where $x$ denotes an input query sequence used to retrieve text documents $d$, incorporating them as additional context during the generation of the target sequence $y$. In this context, $x$ is a question, $q$ while $y$ corresponds to an answer denoted by $a$ in our commonsense QA task. The retrieval process $P_\eta(d|x) \propto exp(Enc(d)^T, Enc(x))$, where $Enc$ functions as an encoder representing both documents and queries. Specifically, $Enc(d)$ stands for the embedding of a document, while $Enc(x)$ signifies a query embedding generated by the encoder. Determining the top-$k$ documents, denoted as $top\text{-}k(p(\cdot|x))$, where the list of $k$ documents $d$ with the highest prior probability $P_\eta(d|x)$ calculated by a similarity function such as cosine similarity. A generator $P_\theta(y_i|x, d, y_{1:i-1})$ generates the current token by considering the context of the preceding $i-1$ tokens $y_{1:i-1}$, the initial input $x$, and a retrieved document $d$.

\subsection{KG-Based Random-Walk Reasoning}
Given a KG with a pair of question, $q$ and question concept, $qc$ as shown in Figure \ref{fig:enter-label}-(a), we seek the node most closely associated with the question concept. Following this initial step of narrowing down the information based on the question concept, we then leverage graph reasoning to extract additional information necessary for formulating a comprehensive answer to the question.

In Figure \ref{fig:model} Step 1, we search for the most similar node aligned with the question concept ($qc$), the term "\textit{liquor}". Employing a pre-trained text encoder denoted as $E$, we encode both the anchor text and the entirety of text associated with nodes within the KG. This process yields embeddings for both the anchor text and all node texts. Subsequently, we calculate the cosine similarity between the embedding of the anchor text and the embeddings of all node texts, establishing similarity scores for each pair.
In this case, $v_{qc}$ is the node \circled{3} in Figure \ref{fig:enter-label}-(a), identified as \textit{alcohol}. Subsequently, we either traverse one hop outbound from the node \textit{alcohol} or find a node that is one hop inbound to \textit{alcohol}. In our case, there are eight one-hop nodes connected to \textit{alcohol}, numbered \circled{1}, \circled{2}, \circled{4}, \circled{5}, \circled{6}, \circled{7}, \circled{9} and \circled{\footnotesize{10}}. Among these, nodes \circled{2} and \circled{9} are inbound, while the others are outbound neighbors. In Figure \ref{fig:model} Step 2, we verbalize triples—structured statements consisting of a subject, predicate, and object—based on their direction relative to node \circled{3}. For example, in the triple \textit{“alcohol, causes, sleep”} node \circled{3}, \textit{alcohol} is the subject, the relationship, \textit{causes} is the predicate, and node \circled{6}, \textit{sleep} is the object. All connections from or to node \circled{3} are similarly transformed into triples, where the relationships between nodes are clearly expressed in natural language. Using the same encoder $E$, we encode both the verbalized triples and the question $q$. We then identify the verbalized triple whose representation is most similar to that of $q$. Through Step 2, we obtain a single verbalized triple structured as a sentence. As the next step, rather than using a similarity function to select the next object node from the current object node \circled{6} (\textit{sleep}), we randomly select $n$. This approach is intended to provide reasoning context following the presentation of the most relevant information just once. We combine these sentences into a question, $q$, and prompt the model to generate an answer based on equation \ref{eq1}.

\section{Experiments}

\subsection{Experimental Setup}
We validate the efficiency of our proposed method in zero-shot setting on Llama 2-Chat \citep{touvron2023llama}, without further training or tuning on the model, to focus on the effect of the KG-based reasoning process. Thus, we assess results across diverse prompt settings, considering different approaches to retrieving and reasoning information.
\paragraph{\textbf{Dataset and Encoder}} We use the CommonsenseQA dataset \citep{talmor2018commonsenseqa} for evaluation emphasizing the need for diverse commonsense knowledge to choose the correct answers. We employed 1,221 data from the validation dataset due to the unavailability of publicly disclosed answers.
To ensure a fair and comprehensive comparison, we choose to employ the ConceptNet KG as the search and reasoning source for both RAG and our proposed method in this experiment. For retrieval purposes, we verbalize the triplets in sentence structures, resulting in 3,423,004 sentences using descriptions of ConceptNet\footnote{\url{https://github.com/commonsense/conceptnet5/wiki/Relations}}. Additionally, we explore a significant web dataset, using Wikipedia as the search source for the "Relevant Information Only" experiment in this study. The Wikipedia embeddings index provided by txtai\footnote{\url{https://huggingface.co/NeuML/txtai-wikipedia\#wikipedia-txtai-embeddings-index}} is employed. The \textit{e5-base} model \citep{wang2022text} is utilized to represent all text documents such as Wikipedia and verbalized ConceptNet triples, as well as questions and question concepts.

\paragraph{\textbf{Evaluation Metrics}} 
We evaluate performance using accuracy as the criterion. In experiments where the output format is incorrect in the prompt, multiple scenarios arise.
Consequently, all experiments are conducted without an explicit output format. Regardless, when analyzing the results, if the correct answer is "B. exercise," valid responses may take the form of "B", "B.", "B,", "exercise" or "X. exercise". Incorrect responses can include options like "A. exercise", where the alphabet is incorrect but the answer is accurate, "B. exercise, C. muscle", when multiple selections are made, or instances where irrelevant statements are presented.

\subsection{Results}
Table \ref{table:main} presents the result of RAG and KG-based random-walk reasoning methods. Our baseline is set as a plain question without additional information. For RAG, we retrieve the top-$k$, where $k = \{1, 2, 3\}$ sentences from verbalized triples within the ConceptNet KG. In the proposed method, we extract sequential triples connecting to an anchor node determined by its similarity score with the question concept. In contrast, "Relevant Information Only" extracts triples exclusively based on text similarity scores with the combined question concept and the question itself. The findings underscore the effectiveness of KG-based reasoning, leveraging information from connected nodes, particularly in scenarios where the number of information sentences remains constant. In our random-walk approach, we prioritized nodes that were physically close to the starting point. For instance, in the "KGI" case when $k$ = 3 (Table \ref{table:main}), nodes 1 through 5 were selected based on their distance of 1 from the start. This criterion was consistently applied across all experiments. Upon careful performance analysis, it becomes apparent that pertinent information situated at the outset or conclusion of the anchor node proves more advantageous than information located in the middle (Graph, $k$ = 2 in Table \ref{table:main}). Our best performance is attained by incorporating comprehensive context, achieved by combining top-1 information of RAG with data derived from KG-based random-walk reasoning.



\begin{table}[hbt!]
\caption{Performance comparison of RAG and KG-based reasoning. For a clear explanation of indicating node location, we assume node 1 is the most similar to the question concept and form the graph sequence as 5 -> 4 -> 1 -> 2 -> 3 ($k$: the number of sentences combined with a question to generate an answer). The highest performance is denoted in bold and the second best results are underlined.}
    \centering
    \scalebox{0.79}{ 
    {
    \begin{tabular}{c|c|c|c}
    \hline\toprule
        
        \textbf{Type}& \textbf{$k$} & \textbf{Node Location} & \textbf{Acc.}\\ \hline\toprule
        Baseline & 0 & - & 0.5684 \\\hline
        \multirow{3}{*}{Relevant Information Only} & 1 & top-1 triple & 0.5864  \\\cline{2-4}
                            & 2 & top-2 triples & 0.5782 \\\cline{2-4}
                            & 3 & top-3 triples & 0.5790 \\\hline
        \multirow{6}{*}{Keyword Relevant Information + Graph Inference (KGI)} & \multirow{2}{*}{1}& 1 -> 2 & 0.5897 \\
                            &  & 4 -> 1 & 0.5766 \\\cline{2-4}
                            & \multirow{3}{*}{2}& (1 -> 2, 2 -> 3) & \underline{0.5913} \\
                            &  & (5 -> 4, 4 -> 1) & 0.5577 \\
                            &  & (4 -> 1, 1 -> 2) & \underline{0.5913} \\\cline{2-4}
                            & \multirow{2}{*}{3}&  (5 -> 4, 4 -> 1, 1 -> 2) & 0.5741  \\
                            &  & (4 -> 1, 1 -> 2, 2 -> 3)  & 0.5741 \\\hline
        Query Relevant Information + Graph Inference (QGI) & 1 + 2 & top-1 + (4 -> 1, 1 -> 2) & \textbf{0.5979} \\\hline

    \hline\toprule
    \end{tabular}
    }
    }
\label{table:main}
\end{table}

\begin{table}[hbt!]
\caption{Performance in situations where the provided information has lower relevance to the question. (R: relevance of information; if "Y," we remain node 1 as the most similar node and randomly select triples from node 1; otherwise, we opt for an unrelated node randomly). The highest performance is denoted in bold and the second best results are underlined.}
    \centering
    \scalebox{0.79}{ 
     {
    \begin{tabular}{c|c|c|c|c}
    \hline\toprule
        
        \textbf{Type}&  \textbf{$k$} & \textbf{R} & \textbf{Node Location} & \textbf{Acc.}\\ \hline\toprule
        Baseline & 0 & - & - & 0.5684 \\\hline
        \multirow{3}{*}{Irrelevant Information Only} & 1 & N & 1 irrelevant triple & 0.5356  \\\cline{2-5}
                            & 2 & N & 2 irrelevant triples & 0.5324 \\\cline{2-5}
                            & 3 & N & 3 irrelevant triples & 0.5397 \\\hline
        \multirow{9}{*}{Graph Inference Only} & \multirow{4}{*}{1}& N & 1 -> 2 & 0.5602 \\
                            &  & N & 4 -> 1 & 0.5602 \\\cline{3-5}
                            & & Y & 1 -> 2 & 0.5479 \\
                            &  & Y & 4 -> 1 & 0.5659 \\\cline{2-5}
                            & \multirow{3}{*}{2} & N & (1 -> 2, 2 -> 3) & \underline{0.5717}   \\
                            &  & N & (5 -> 4, 4 -> 1) & 0.5635 \\
                            &  & N & (4 -> 1, 1 -> 2) & 0.5561 \\\cline{2-5}
                            & \multirow{2}{*}{3}& N & (5 -> 4, 4 -> 1, 1 -> 2) & 0.5667  \\
                            &  & N & (4 -> 1, 1 -> 2, 2 -> 3)  & \textbf{0.5758} \\\hline

    \hline\toprule
    \end{tabular}
    }
    }
\label{table:bottom}
\end{table}

In Table \ref{table:bottom}, we investigate the setting in which the given information is less relevant to the question and its concept. It is crucial observation that even when opting for a less related anchor node and executing a random-walk to obtain $k$ sentences, there is an observed enhancement in performance (Graph, $k$ = 2 and 3 in Table \ref{table:bottom}). This suggests that reasoning abilities, such as connection of node relationships, contribute to problem-solving. Conversely, inputting less relevant information without a coherent flow or reasoning in "Irrelevant Information Only" proves to be ineffective in performance. 

We explore diverse prompt configurations and retrieval source datasets, detailed in Table \ref{table:ablation}, to assess the impact of incorporating extra information into the question. Our findings reveal that using a substantial web dataset, Wikipedia, as a source dataset does not enhance task performance. It is noteworthy that both Relevant Information and Graph scenarios experience decreased performance when incorporating more than three information sentences. The order of prompting is crucial, showing superior performance when retrieved documents precede the question rather than following it. Additionally, the direction of reasoning in the graph is essential, as evidenced by reduced performance when ignoring edge direction.

\begin{table}[hbt!]
\caption{Evaluating performance variations across various prompt configurations (Prompt Engineering: order of prompt, direction of reasoning information).}
    \centering{
    \scalebox{0.79}{ 
    \begin{tabular}{c|c|c|c}
    \hline\toprule
        
        \textbf{Type}& \textbf{$k$} & \textbf{Prompt Engineering} & \textbf{Acc.}\\ \hline\toprule
        \multirow{3}{*}{Relevant Information Only with the ConceptNet Graph} & \multirow{2}{*}{1}& documents -> question & 0.5864  \\\cline{3-4}
                            &  & question -> documents & 0.5455 \\\cline{2-4}
                            & 4 & documents -> question & 0.5635 \\\hline
        \multirow{3}{*}{Relevant Information Only with Wikipedia} & 1 & documents -> question & 0.5455  \\\cline{2-4}
        & 2 & documents -> question & 0.5504  \\\cline{2-4}
         & 3 & documents -> question & 0.5463  \\\hline
        \multirow{3}{*}{Graph} & \multirow{2}{*}{2}&  irregular direction (1 -> 2, 4 -> 1) & 0.5807  \\\cline{3-4}
                            &   &  regular direction (4 -> 1, 1 -> 2) & 0.5913 \\\cline{2-4}
                            & 4  & (5 -> 4, 4 -> 1, 1 -> 2, 2 -> 3) & 0.5717 \\\hline

    \hline\toprule
    \end{tabular}
    }
    }
\label{table:ablation}
\end{table}
 

\section{Conclusion}
Our experiments were designed to investigate the impact of information retrieval exemplified by Relevant Information, and the reasoning process represented by KG-based random-walk, on improving commonsense QA. Consequently, delivering outcomes inferred through the proposed method generally led to better results than supplying relevant information in Relevant Information. Additionally, the experimental results matched our expectation that performance would be most improved when utilizing both information extracted by embedding matching and graph reasoning. However, we also obtained unexpected results where performance improved when providing irrelevant information through graph reasoning. We analyze that these results indicate providing a reasoning process can enhance the performance of commonsense QA.

The main limitation of this paper is that it is restricted to the commonsense QA task. However, since this task requires both reasoning ability and information with specific information, our experiments have empirically proven which method is more effective. This suggests that focusing on enhancing reasoning capabilities could be beneficial for improving commonsense QA performance in the future.

\bibliographystyle{ACM-Reference-Format}
\bibliography{neurips_2024_calm}

\end{document}